\documentclass[sigconf, nohyperref]{acmart}

\usepackage{booktabs}
\usepackage{makecell}

\setcopyright{rightsretained}

\acmConference[ICVGIP'25]{16th Indian Conference on Computer Vision, Graphics and Image Processing}{December 2025}{Mandi, India}
\acmYear{2025}
\copyrightyear{2025}

\acmPrice{15.00}

\begin{document}
\title{Light-Field Dataset for Disparity Based Depth Estimation}

\author{Suresh Nehra}
\affiliation{%
  \institution{Indian Institute of Technology Kharagpur}
  \country{India}
}
\author{Aupendu Kar}
\authornote{Work done while Aupendu Kar was at the Indian Institute of Technology Kharagpur}
\affiliation{%
  \institution{Dolby Laboratories, Inc}
  \country{India}
}
\author{Jayanta Mukhopadhyay}
\affiliation{%
  \institution{Indian Institute of Technology Kharagpur}
  \country{India}
}
\author{Prabir Kumar Biswas}
\affiliation{%
  \institution{Indian Institute of Technology Kharagpur}
  \country{India}
}

\renewcommand{\shortauthors}{}

\begin{abstract}
A Light Field (LF) camera consists of an additional two-dimensional array of micro-lenses placed between the main lens and sensor, compared to a conventional camera. The sensor pixels under each micro-lens receive light from a sub-aperture of the main lens. This enables the image sensor to capture both spatial information and the angular resolution of a scene point.  This additional angular information is used to estimate the depth of a 3-D scene.  The continuum of virtual viewpoints in light field data enables efficient depth estimation using Epipolar Line Images (EPIs)  with robust occlusion handling. However, the trade-off between angular information and spatial information is very critical and depends on the focal position of the camera. To design, develop, implement, and test novel disparity-based light field depth estimation algorithms, the availability of suitable light field image datasets is essential. In this paper, a publicly available light field image dataset is introduced and thoroughly described. We have also demonstrated the effect of focal position on the disparity of a 3-D point as well as the shortcomings of the currently available light field dataset. The proposed dataset contains 285 light field images captured using a Lytro Illum LF camera and 13 synthetic LF images. The proposed dataset also comprises a synthetic dataset with similar disparity characteristics to those of a real light field camera. A real and synthetic stereo light field dataset is also created by using a mechanical gantry system and Blender. The dataset is available at \href{https://github.com/aupendu/light-field-dataset}{https://github.com/aupendu/light-field-dataset}.
\end{abstract}

%
%



\maketitle

\section{Introduction}
\label{sec:introduction}

\begin{figure}[!htp]
\centering
\includegraphics[width = 8cm]{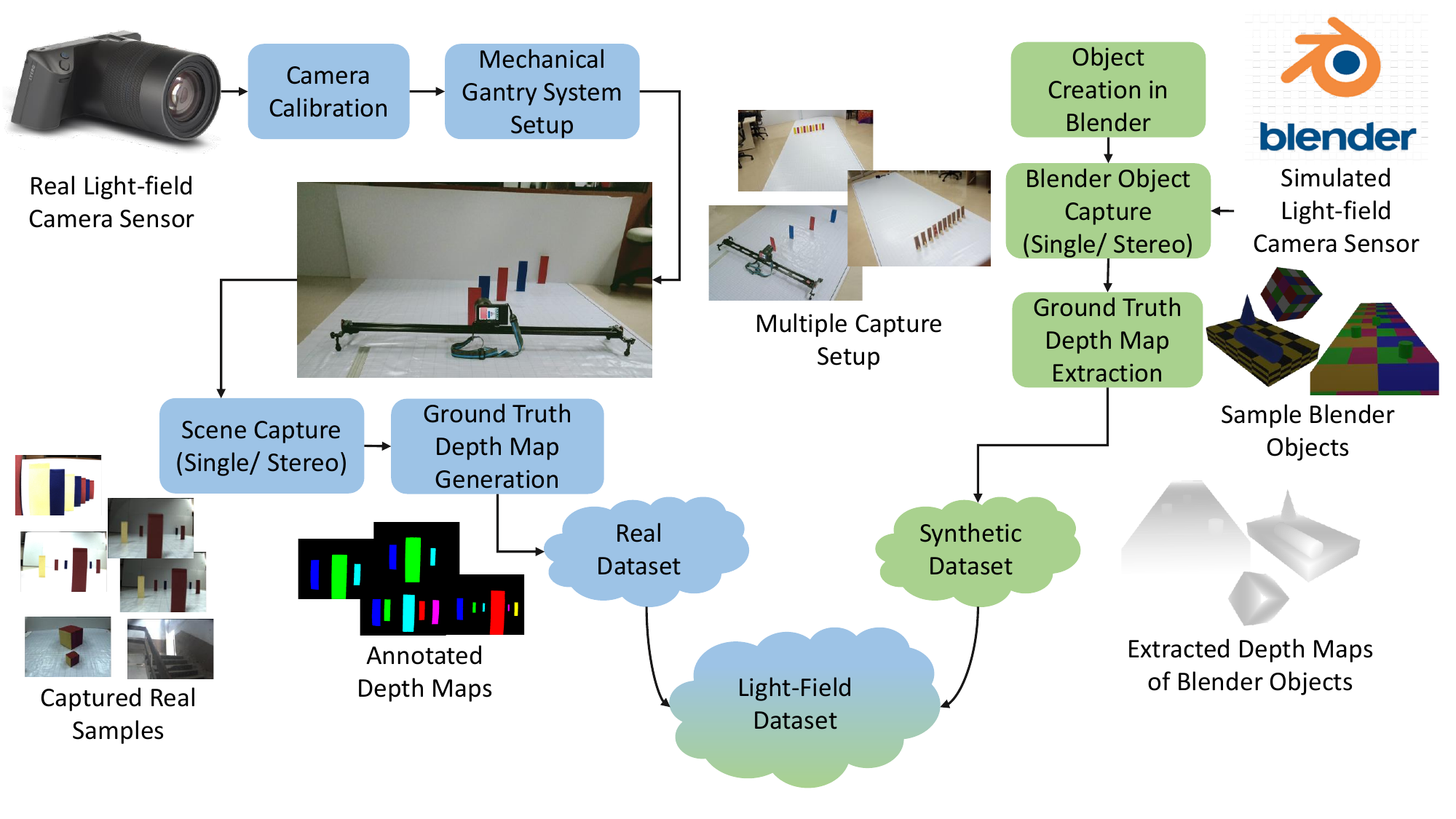}
\vspace{-10pt}
\caption{Graphical description of our proposed dataset}
\label{GA}
\vspace{-10pt}
\end{figure}
The basic idea of capturing Light-Field (LF) data originates from the concept of integral photography introduced by Lippmann~\cite{Lippmann}. In general, LF images can be recorded by using either multi-camera arrays or more recent single-camera solutions. The earlier light-field acquisition was very challenging and expensive. Like Lytro Illum and Raytrix, the latter acquisition system utilizes additional optical elements, consisting of a micro-lens array placed between the main lens and the sensor array. LF photography represents an imaging method that not only captures the light intensity but also the direction of the incoming light rays. This capability to capture additional directional information enables rich features, allowing for a more realistic, interactive, and immersive user experience. The additional directional information of light rays brings a whole range of new possibilities for LF image post-processing. This directional information in LF also helps to improve the performance in different computer vision applications such as refocusing~\cite{jayaweera2020multi}, depth estimation~\cite{EPINET, li2018robust, jeon2018depth, nehra2021disparity}, image segmentation~\cite{khan2019view}, image editing~\cite{shon2016spatio, jarabo2014people}, virtual reality~\cite{yu2017light}, image super-resolution~\cite{resLF, LFCNN2015, LFCNN2017, gul2018spatial, LFNet, LF-DCNN}. 
\begin{figure*}[!htp]
\centering
\includegraphics[width = 16.0cm]{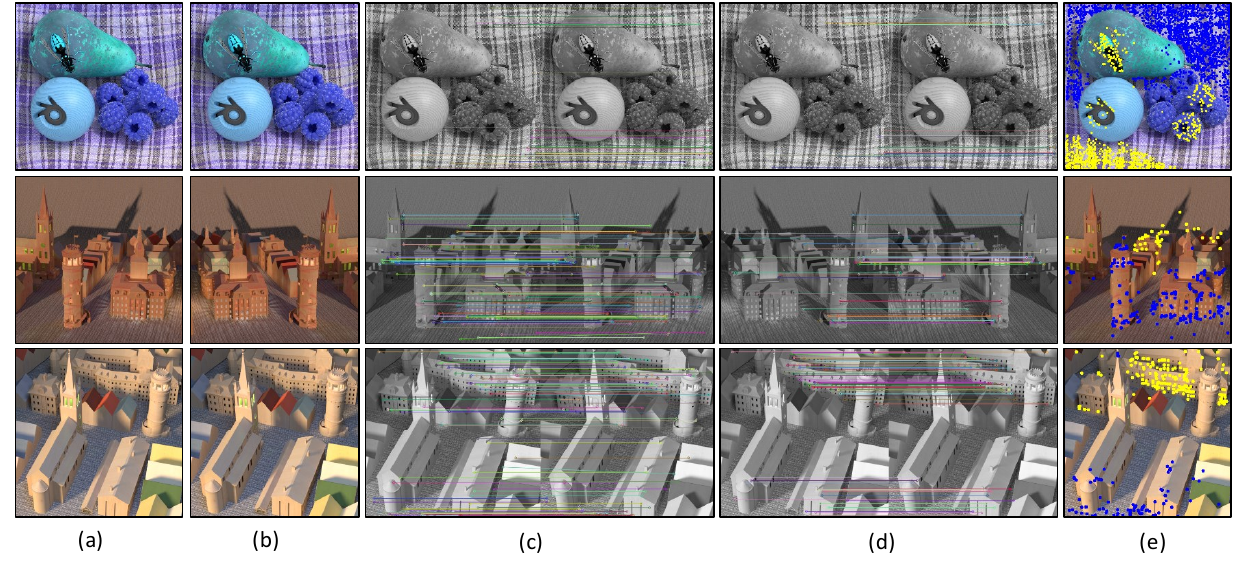}
\vspace{-10pt}
\caption{Heidelberg New and Old Synthetic Benchmark Dataset. Top to bottom -\textit{Still life, Tower} and \textit{Town}. (a) left view, (b) right view, (c) point correspondences, (d) refined point correspondences, (e) Color markers where the blue color represents key points with a negative disparity and the yellow color represents the positive disparity.}
\vspace{-10pt}
\label{fig: other_synthetic_keypoints}
\end{figure*}

Among these, depth estimation is one of the most active areas of research. Epipolar Line Images (EPIs) are used in light-field depth estimation, where epipolar geometry is employed to estimate depth. Due to the dense sampling in a light-field, it provides a continuum of viewpoints. With the help of EPIs, we can estimate structure from motion~\cite{bolles1987epipolar} or depth from slopes of EPIs in an efficient manner. However, for estimating depth from EPIs in LF images, LF data must hold two conditions. First, an inverse relationship between depth and disparity of scene points. Second, there should be a sufficient disparity difference between objects at different distances to resolve depth among the points. In publicly available light-field datasets, these aspects have not been properly taken care of, and due to this, these datasets are unsuitable for disparity-based LF depth estimation methods. Therefore, it is important to understand this behavior while capturing data from an LF camera. We have conducted experiments to study this behavior. This paper also introduces a new publicly available LF image dataset, which was acquired using a Lytro Illum camera.  We also introduced a synthetic LF dataset, which has characteristics similar to those of a real single and stereo LF camera. The proposed dataset significantly enhances the possibilities of addressing the emerging challenges of using the existing LF dataset for disparity-based light-field depth estimation methods.

The main contributions of this paper are as follows.
\begin{itemize}
    \item Presented a comparative study on the depth disparity variation in real scenes against the publicly available synthetic scenes.
    \item Analyzed a study on the dependence of angular information on focal distance variation in real light-field cameras.
    \item Introduced a real light-field dataset using Lytro Illum, suitable for disparity-based light-field depth estimation methods. For a subset of scenes, ground truth for depth estimation is also provided.
    \item It also consists of stereo LF data captured using Lytro Illum and a mechanical gantry, which can be used to test various stereo-LF depth estimation algorithms.
    \item Our real dataset exhibits a wide range of variations, including different lighting conditions and ISO settings, which can be utilized to evaluate the performance of various algorithms, such as low-light LF image enhancement, LF image denoising, LF image segmentation, and LF super-resolution. 
    \item Introduced a synthetic light-field dataset using Blender, having characteristics similar to a real light-field camera. 
\end{itemize}

\begin{figure*}[!htp]
\centering
\includegraphics[width = 16.0cm]{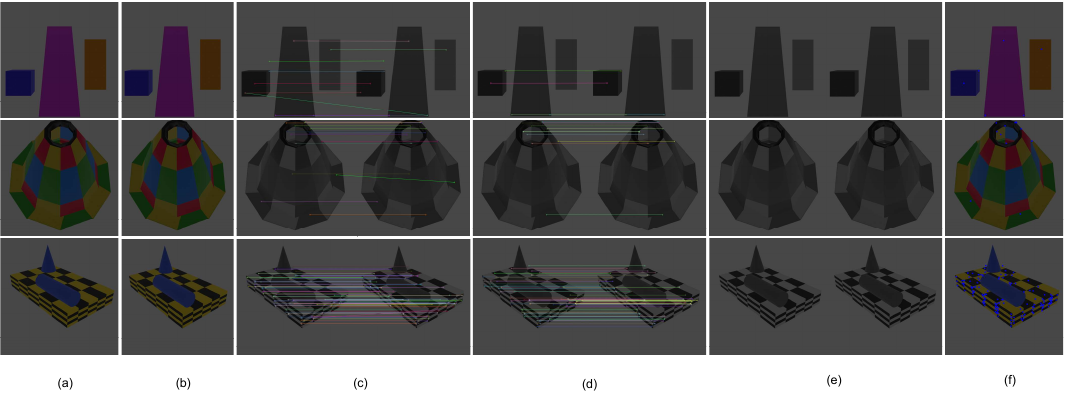}
\vspace{-10pt}
\caption{Our Synthetic Data-set. Top to bottom -\textit{Geometric, Cone} and \textit{Cone and cylinder.} (a) left view, (b) right view, (c) point correspondences, (d) refined point correspondences, (e) correspondences with positive disparity, (f) Color markers, blue color for key points having negative disparity, yellow color for the positive disparity.}
\vspace{-10pt}
\label{fig: our_synthetic_keypoints}
\end{figure*}

\section{Related Work}

Owing to the potential capabilities from additional directional information of the light, a large variety of fields have tried using LF as input rather than a single image, introducing a series of datasets that can be classified into real-world LF captured by a camera array, a gantry, or a plenoptic camera, and synthetic LF by Blender or other software. 

For saliency detection, Li \textit{et al.}~\cite{li2014saliency} provided a dataset captured using Lytro, consisting of $60$ indoor and $40$ outdoor LF images with a ground truth saliency map. Zhang \textit{et al.}~\cite{zhang2020light} dataset consisting of $640$ LF images captured using Lytro Illum with more variations in illuminance, scale, and position. Piao \textit{et al.} \cite{piao2020dut} presented an LF dataset with $4204$ images captured using Lytro Illum. 
For material recognition, Wang \textit{et al.}~\cite{wang20164d} provided an LF dataset consisting of images from $12$ material categories, each with $100$ samples with labeled categories. This dataset was captured using Lytro Illum.
For object recognition, Xu \textit{et al.}~\cite{xu2019transcut2} provided $49$ LF images with seven transparent objects in seven different kinds of background scenes. 
For semantic segmentation, Jia \textit{et al.}~\cite{jia2021semantic} provided $400$ real-world scenes captured using Raytrix-R8 and Lytro Illum, with annotations for three objects. Sheng \textit{et al.}~\cite{sheng2022urbanlf} provided $824$ real LF images using Lytro Illum and $250$ synthetic images using Blender. This dataset has annotations for $14$ semantic classes in urban scenes.

For light-field super-resolution, Rerabek \textit{et al.}~\cite{EPFL} presented an LF dataset consisting of $118$ images of $10$ different categories, captured using Lytro Illum. Raj \textit{et al.}~\cite{STFlytro} provided an LF dataset with $353$ images captured using Lytro Illum.  Pendu \textit{et al.}~\cite{INRIA} released a dataset consisting of $109$ images captured using Lytro ($63$ samples) and Lytro Illum ($46$ samples). 
For light-field deblurring, Lumentut \textit{et al.}~\cite{lumentut2019fast} provided $200$ real and $220$ synthetic LF images using Lytro Illum and UnrealCV, along with a camera motion model to synthesize blurry training samples.
For all-in-focus restoration, Ruan \textit{et al.}~\cite{ruan2021aifnet} presented a dataset with $839$ LF images using Lytro Illum.
For quality assessment, Kiran \textit{et al.}~\cite{kiran2017towards} provided $5$ real and $9$ synthetic scenes using a Canon camera and motorized gantry. 

For view synthesis, Kalantari \textit{et al.}~\cite{30scenes} captured $130$ LF samples using Lytro Illum. Srinivasan \textit{et al.}~\cite{srinivasan2017learning} provided $3343$ LF samples using Lytro Illum. It provides images of flowers and plants with complex occlusions.
For intrinsic decomposition, Alperovich \textit{et al.}~\cite{alperovich2018light} provided $175$ synthetic scenes rendered using Blender. It also provides a custom LF generator that can synthesize LF images with diffuse and specular intrinsic components. 

For light-field depth estimation, Wanner \textit{et al.}~\cite{wanner2013datasets} presented a light-field dataset consisting of $6$ real scenes and $13$ synthetic scenes. The real scenes are captured using a Nikon D800 camera, whereas the synthetic scenes are rendered using Blender with densely sampled light fields. Honauer \textit{et al.}~\cite{honauer2016dataset} provided a dataset with $28$ synthetic scenes with ground-truth, rendered using Blender. Heber \textit{et al.}~\cite{heber2016convolutional} presented a synthetic LF dataset using a raytracing software POV-Ray with ground truth depth fields. This dataset consists of $25$ synthetic light-field images. Zhang \textit{et al.}~\cite{zhang2021joint} provided light-field data with $720$ real images from six different scenes using Lytro Illum. The ground truth depth of these scenes was obtained using an RGB-D camera. Shi \textit{et al.}~\cite{shi2019framework} presented a light-field dataset with $235$ real and $13$ synthetic light-field images. The real light-field images were captured using Lytro Illum, and synthetic scenes were rendered using Blender.

\section{Light-field Image Dataset}
\subsection{Limitations in Publicly Available Datasets}
\label{subsec:lf_data_capture}

Publicly available datasets are not suitable for disparity-based depth estimation. In the case of a real light-field camera, the virtual viewpoints lie on a two-dimensional plane. Due to this, we have an intrinsic inverse relationship between depth and disparity. These publicly available synthetic datasets are rendered using 3-D rendering software. These light-field images are rendered using spherical \textit{uv} and \textit{st} planes \cite{chen1995quicktime}, which involve the rotation and translation of viewpoints. Consequently, these synthetic datasets do not hold the inverse relationship between depth and disparity.  

Another problem with the existing light-field dataset is the variation in disparity. There should be enough disparity difference between objects at different distances to resolve the depth among the points. In the case of a real light-field camera, the disparity distribution changes with a change in focal position. If the focal position is not set properly, it may result in a disparity of close to zero or a constant disparity for all SAI image pixels.  Therefore, it is important to understand this behavior while capturing data from a light-field camera. We have conducted experiments to investigate this behavior and discuss the results later in this paper. In publicly available real-world scenes, this aspect has not been properly addressed, and as a result, these publicly available real-world datasets are unsuitable for disparity-based light-field depth estimation methods. 

\begin{figure*}
\centering
\includegraphics[width = 16.0cm]{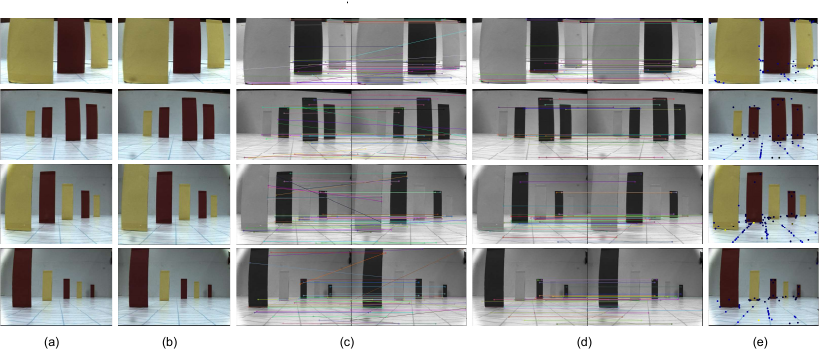}
\vspace{-10pt}
\caption{Our Real Data-set. Top to bottom -\textit{ Scene 1, Scene 2, Scene 3} and  \textit{Scene 4.} (a) left view, (b) right view, (c) point correspondences, (d) refined point correspondences, (e) correspondences with positive disparity, (f) Color markers, blue color for key points having negative disparity, yellow color for the positive disparity.}
\label{fig: our_real_keypoints}
\vspace{-10pt}
\end{figure*}

In the case of micro-lens-based LF cameras, virtual viewpoints lie on a two-dimensional plane with equally spaced virtual camera centers. Due to these inherent characteristics, there is a translation of virtual viewpoints, and it induces an inverse relationship between disparity and depth. This relationship between disparity and depth is exploited to estimate the depth of a scene. However, in the publicly available synthetic datasets, light-field scenes are rendered with spherical planes \textit{uv} and \textit{st}~\cite{levoy1996light}. Therefore, these datasets involve both rotation and translation of viewpoints. When rotation among the views is involved, disparity does not maintain a consistent relationship with depth. Due to the rotation, we get positive and negative disparity values for different scene points. A smaller disparity may be possible for far-away objects and a larger disparity for nearby objects. It renders depth ordering independent of disparity, making it difficult to estimate depth from disparity. 
\begin{figure*}
\centering
\includegraphics[width = 16.0cm]{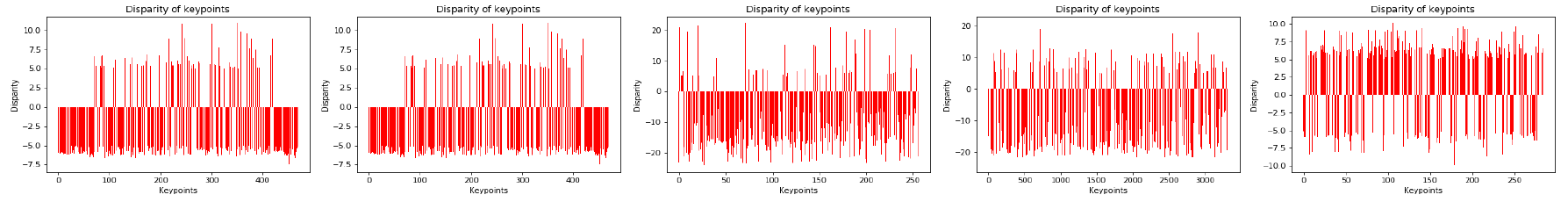}
\vspace{-10pt}
\caption{Heidelberg New and Old Benchmark Datasets: disparity distribution over the key-points when disparity is greater than 2. (a) \textit{Buddha}, (b) \textit{Buddha2}, (c) \textit{Still Life}, (d) \textit{Tower}, (e) \textit{Town.}}
\label{fig: other_synthetic_histo}
\vspace{-10pt}
\end{figure*}
\begin{figure*}
\centering
\includegraphics[width=16.0cm]{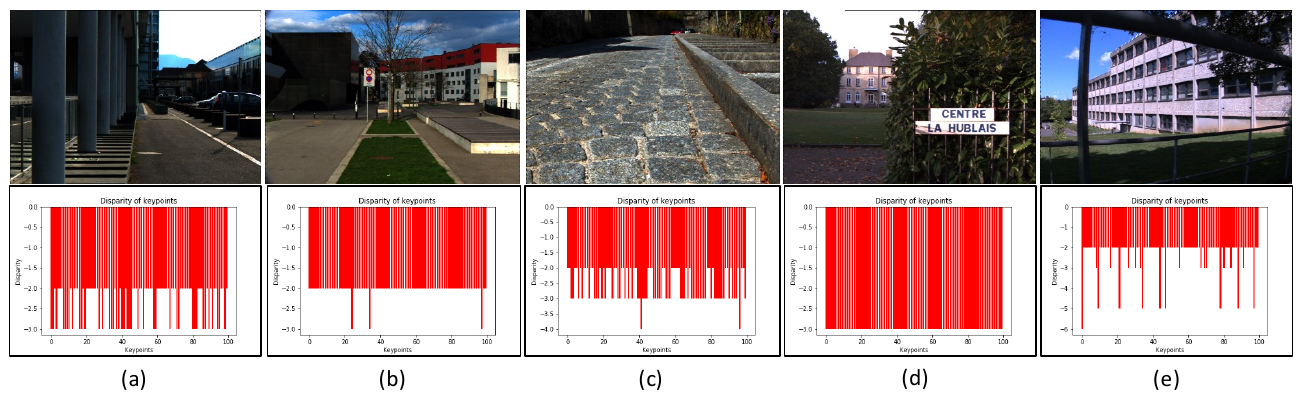}
\vspace{-10pt}
\caption{EPFL and INRIA Lytro Dataset. Top to bottom - Sub-aperture image and disparity variation. (a) Pillars, (b) Red \& White Building, (c) Paved from \textit{EPFL}, (d) Hublias, (e) Building from \textit{INRIA Lytro}.}
\label{other_real_keypoints}
\vspace{-10pt}
\end{figure*}

\subsubsection{Depth-disparity Relationship}
\label{Depth-disparity Relationship}

We compared the publicly available dataset and our proposed dataset to analyze the depth-disparity relationship. We consider the middle row of viewpoints of an LF image and take the left-most SAI \textit{$I_{U/2;0}(s,t)$} and right-most SAI \textit{$I_{U/2;V}(s,t)$} to establish a stereo pair for disparity analysis. For a particular key-point, the disparity in the x-direction is calculated as $(x_R-x_L)$, where $x_L$,  $x_R$ are the x-coordinates of a key point in the left and the right image. Similarly, we can define disparity along the y-direction. The SIFT algorithm is used to detect key points in both the left and right views. For stereo correspondences, K-Nearest Neighbour brute-force matching is used to obtain the best k-nearest matches. After that, we applied a ratio test to get the best-matched key points. Figure~\ref{fig: other_synthetic_keypoints} shows a key-point analysis on the Old and New Heidelberg Synthetic Benchmark Dataset. Figures~\ref {fig: our_synthetic_keypoints} and~\ref {fig: our_real_keypoints} show the key-point analysis on our proposed synthetic and real datasets.

\begin{figure*}
\centering
\includegraphics[height = 2.7cm]{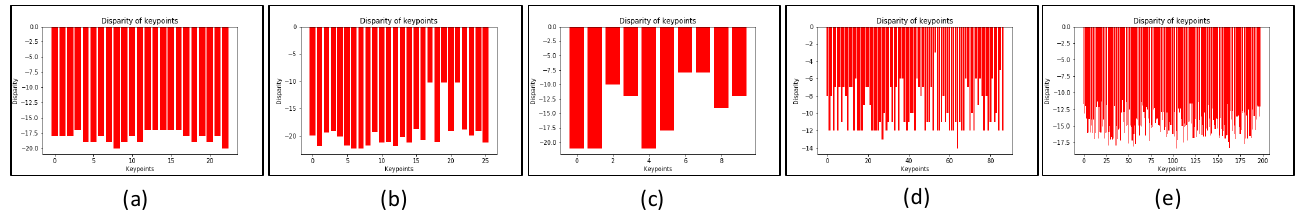}
\vspace{-10pt}
\caption{Our Synthetic Data-set - disparity distribution over the key-points when disparity is greater than 2. (a) \textit{Cube}, (b) \textit{Sphere}, (c) \textit{Geometric}, (d) \textit{Cone}, (e) \textit{Cone and cylinder}}
\label{fig: our_synthetic_histo}
\vspace{-10pt}
\end{figure*} 
\begin{figure*}
\centering
\includegraphics[height = 2.7cm]{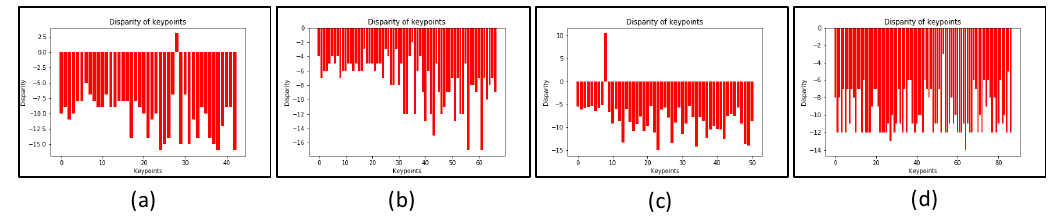}
\vspace{-10pt}
\caption{Our Real Data-set - disparity distribution over the key-points, when disparity is greater than 2. (a) \textit{Scene 1}, (b) \textit{Scene 2}, (c) \textit{Scene 3}, (d) \textit{Scene 4}.}
\label{fig: our_real_histo}
\vspace{-10pt}
\end{figure*}

The stereo correspondences are shown in Figure~\ref{fig: other_synthetic_keypoints} (c). We observe that there are numerous key-point correspondences with outliers. In some datasets, we still have false stereo correspondences due to the repeating patterns. Therefore, those key points need to be filtered out for better disparity analysis. We use two criteria to find good candidate pairs for stereo correspondences. The first method involves applying a threshold to the disparity value in the $x$-direction, while the second method imposes a threshold on the disparity value in the $y$-direction. By observing different light-field datasets, we can determine the maximum possible disparity value. We use this maximum possible disparity value and also set a minimum disparity value to get good candidate pairs of key points. We empirically found the maximum possible disparity value to be $25$ pixels, and we set the minimum disparity value to two pixels in the x-direction. Additionally, in our experiment, the right view is horizontally displaced with respect to the left view. Therefore, for a particular scene point, the disparity should be in the $x$-direction and a minimal disparity in the $y$-direction. We set the maximum allowable disparity value along the y-direction to $2$ pixels for a candidate key point. After applying these two criteria, we reduce the number of candidate point correspondences for our disparity analysis, and the refined point correspondences are shown in Figure~\ref{fig: other_synthetic_keypoints} (d).

Figure~\ref{fig: other_synthetic_keypoints} (e) shows the circular markers around the candidate key-points in the left view to showcase the positive and negative disparity. The blue color represents a negative disparity, whereas the yellow color represents a positive disparity. In the case of translation between two viewpoints, the disparity should always be negative with reference to the left view. However, we observe from Figure~\ref{fig: other_synthetic_keypoints} (e) that the observed disparities of the candidate key-points are both positive and negative with reference to the left view. Figure~\ref{fig: our_synthetic_keypoints} (e) and Figure~\ref{fig: our_real_keypoints} (e) show the disparity variation in our proposed synthetic and real datasets. We clearly observe from those figures that disparity values for our datasets are negative for all the candidate key-points, following the intrinsic property of real light-field camera images.

We also analyzed the disparity distribution over the candidate key points by plotting disparity values against the key points. Figure~\ref{fig: other_synthetic_histo} shows the disparity variations for the publicly available Old and New Heidelberg Synthetic Benchmark Datasets. Figure~\ref{fig: our_synthetic_histo} and Figure~\ref{fig: our_real_histo} show the disparity variations of our proposed synthetic and real light-field datasets. We observed that publicly available synthetic datasets exhibit both positive and negative disparities and do not exhibit inverse relationships between disparity and depth. Therefore, it does not truly capture the characteristics of virtual viewpoints of a real light-field camera. On the other hand, our proposed synthetic and real datasets have only negative disparity. It also maintains the depth order, and the nearer object has a larger disparity than the farther-away object. However, there are very few points in our real dataset where we observe positive disparity. This is mainly due to improper correspondence matching as real light-field images have low spatial resolution $(434\times625)$.

\subsubsection{Focus-disparity Relationship}

In micro-lens-based LF cameras like Lytro Illum, there is a trade-off between the angular information and the spatial information depending on the focal position. The change in focal position changes the imaging plane of the main lens, which changes the baseline among the virtual viewpoints. This change in baseline affects the angular information present in an LF image. If the virtual baseline among different viewpoints is larger, we will have more angular information, or we will expect more disparity variation under each micro-lens for a scene point among different virtual viewpoints. However, we obtain a defocused image for that scene point, which ultimately reduces the spatial information of that scene point. On the other hand, when the virtual baseline among different viewpoints is close to zero, all the pixels under each micro-lens would receive the light rays from a single scene point. This enhances the spatial resolution but reduces the angular information. The distribution of the angular information (disparity variation) under each micro-lens changes with a change in focal distance. This trade-off affects the disparity information captured in an LF image. Therefore, a balance between spatial and angular information should be maintained by selecting the appropriate focal position.

\begin{figure}
\centering
\includegraphics[width = 5.5cm]{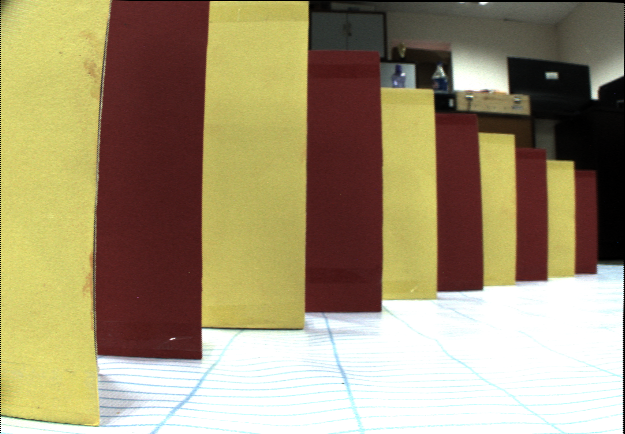}
\vspace{-10pt}
\caption{Center SAI of a scene for study of focal position variation. Ten planar objects are placed at a known distance from the camera using a flex grid. The leftmost object (yellow colored) is placed 20 cm from the camera, and then subsequent objects are placed at a spacing of 10 cm.}
\label{focusVariation}
\vspace{-10pt}
\end{figure}

\begin{figure}
\centering
\includegraphics[width  = 7.3cm]{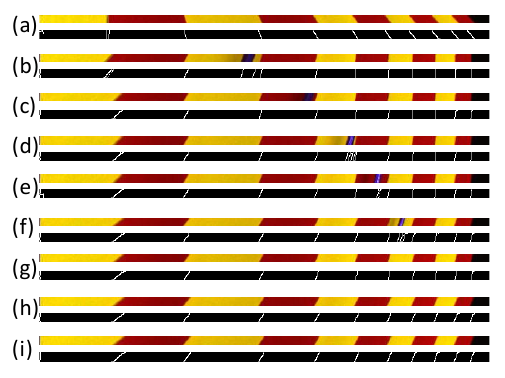}
\vspace{-10pt}
\caption{Horizontal Epipolar line Images (EPIs) are constructed from the central row of horizontal viewpoints to study the variation of End-to-End Disparity (EED)  with change in focal position. RGB EPI and edge EPI with (a) Focus at 20cm, (b) Focus at 30cm, (c) Focus at 40cm, (d) Focus at 50cm, (e) Focus at 60cm, (f) Focus at 70cm, (g) Focus at 80cm, (h ) Focus at 90cm, (i) Focus at 100cm.}
\label{focusVariationEPI}
\vspace{-10pt}
\end{figure}

\begin{figure}
    \centering
    \includegraphics[width = 7.3cm]{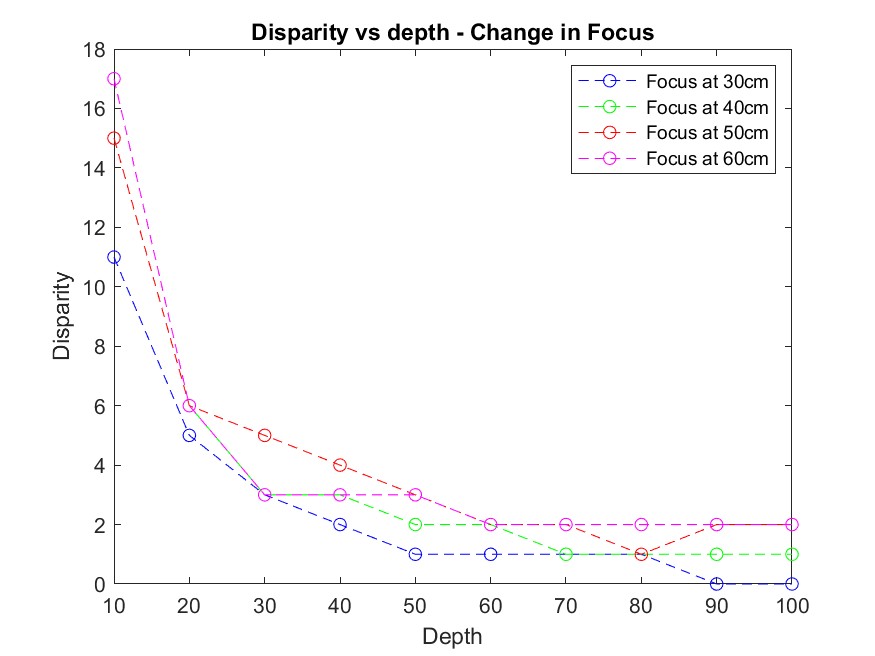}
    \vspace{-10pt}
    \caption{End-to-End Disparity (EED) vs Depth for Focal Positions. Objects are placed from 20 cm to 60 cm, and each planar object is placed 10 cm apart from the others.}
   \label{focus_change_20_60}
   \vspace{-10pt}
\end{figure}

To demonstrate this experimentally, we have created an experimental setup as shown in Figure~\ref{focusVariation}. In the setup, ten planar objects are placed on a flex paper with grid markings at known distances from the camera. Multiple LF images are captured at different focal distances without changing the scene to analyze the disparity variation with varying focal distances. For disparity variation, a horizontal EPI of the central row \textit{$E_{(V/2;T/2)}(s, u)$} is constructed as shown in Figure~\ref{focusVariationEPI}.  

\begin{figure}
    \centering
    \includegraphics[width = 7.3cm]{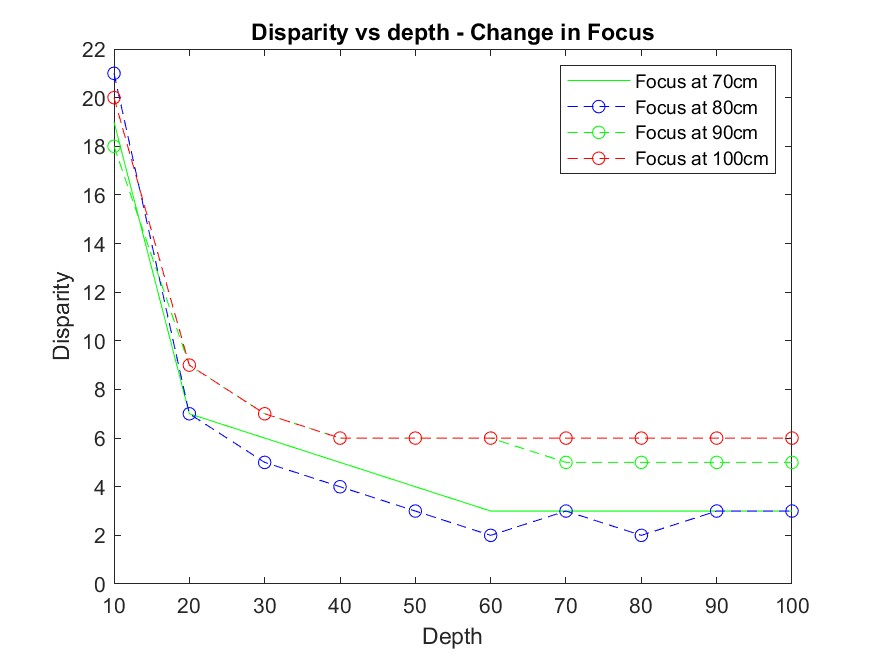}
    \vspace{-10pt}
\caption{End-to-End Disparity (EED) vs Depth for Focal Positions. Objects are placed from 70 cm to 100 cm, and each planar object is placed 10 cm apart from the others.}
\label{focus_change_70_100}
\vspace{-10pt}
\end{figure}

A line in EPI corresponds to an edge point of an object. Therefore, we can estimate the End-to-End Disparity (EED) of each object by measuring the change in abscissa of the corresponding line in an EPI. An edge detector is used to obtain these lines in an EPI, and then EED is computed for each object at different distances. Horizontal EPIs are constructed for different focal distances, as shown in Figure~\ref {focusVariationEPI} (a)-(e). We observe from the figure that the slopes of these lines change with a change in focal distance, indicating that the disparity of an object at the same distance also changes with focal distance. To analyze this behavior of disparity with focal distance, we compute EED at different distances with respect to focal distance, and this variation of EED is shown in Figure~\ref{focus_change_20_60}, Figure~\ref{focus_change_70_100}. We observe from Figure~\ref{focus_change_20_60} and Figure~\ref{focus_change_70_100} that disparity variation increases as the focal distance increases up to a certain focal distance, which is 80 cm in our case. After that, it starts decreasing. 
However, as the scene distance increases further, the EED variation reduces. This behavior is expected because at higher distances for multi-view systems, depth resolution is reduced. Due to the small virtual baseline of a micro-lens-based camera like Lytro Illum, the disparity begins to saturate when we move away from the hyperfocal distance (approximately 100cm). Due to this, the depth-resolving power of a light-field camera decreases, limiting its operational range for depth estimation. For Lytro Illum, this operating range for depth estimation is around 100cm.

\begin{figure}
    \centering
    \includegraphics[width=6cm]{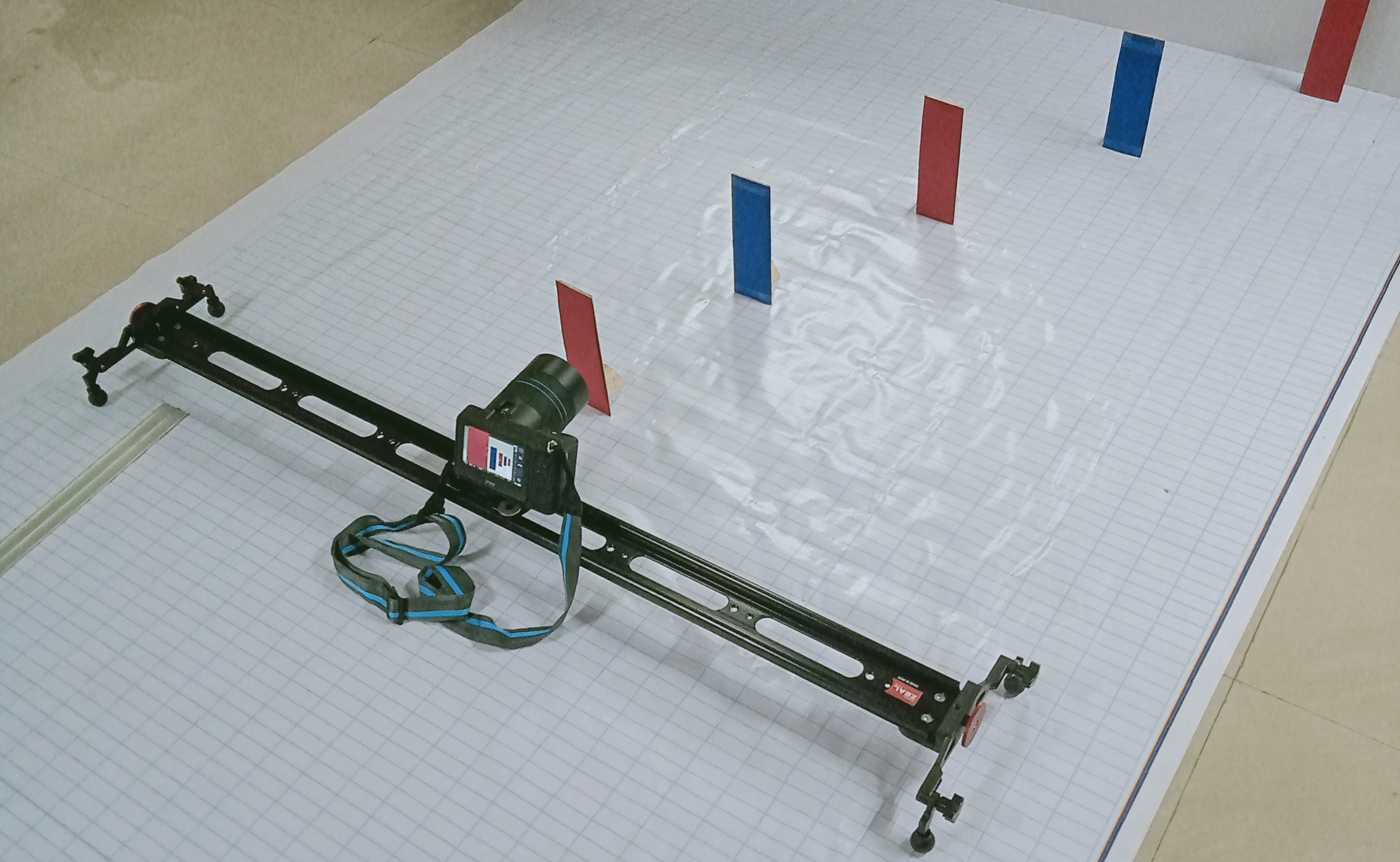}
    \vspace{-10pt}
\caption{Real Camera Setup.}
\label{camerasetup}
\vspace{-10pt}
\end{figure}
\begin{figure}[]
    \centering
    \includegraphics[width=8cm, trim=0cm 4.2cm 0cm 0cm, clip]{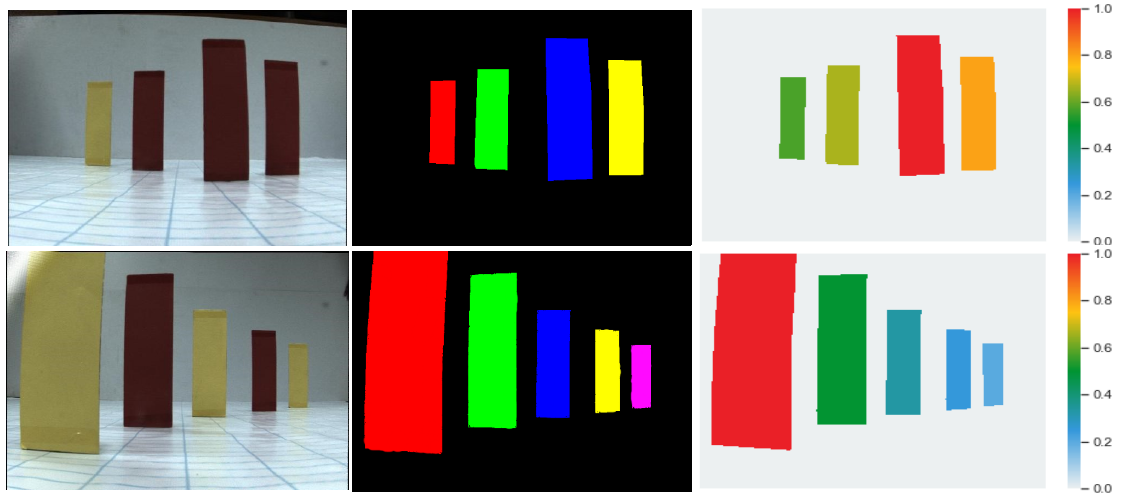}
    \vspace{-10pt}
\caption{Real scenes using  Lytro Illum camera with foreground and depth map.}
\label{realScene1}
\vspace{-10pt}
\end{figure}

\begin{figure}[]
    \centering
    \includegraphics[width=9cm, trim=0cm 4.5cm 0cm 0cm, clip]{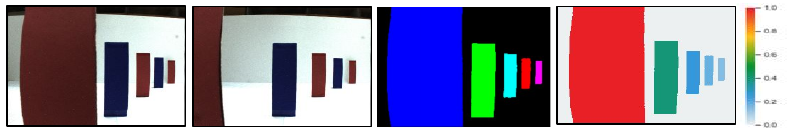}
    \vspace{-10pt}
\caption{Real stereo scene using  Lytro Illum camera with foreground and depth map.}
\label{realStereo}
\vspace{-10pt}
\end{figure}
\begin{figure}[]
    \centering
    \includegraphics[width=8.0cm, trim=5.7cm 0cm 5.7cm 0cm, clip]{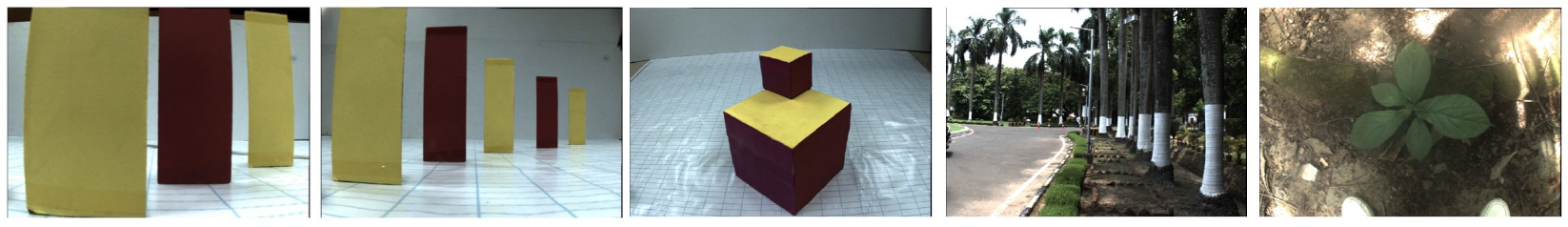}
    \vspace{-10pt}
\caption{Real scenes using a single Lytro Illum camera.}
\label{realScene}
\vspace{-10pt}
\end{figure}

This dependence of disparity on focal distance and operating range has not been accounted for in most publicly available real-world LF datasets, such as INRIA \cite{INRIA} and EPFL \cite{EPFL}. In these datasets, the subjects are mostly positioned far away from the camera, or the focal distance is not suitable for capturing a proper disparity variation, which is necessary for disparity-based LF depth estimation methods. To validate this, we have employed the stereo disparity approach presented in Section \ref{Depth-disparity Relationship} and computed the disparity of key points in the leftmost and rightmost virtual viewpoints of the central row of virtual viewpoints. Disparity variation of these key-points is shown in Figure \ref{other_real_keypoints}. We observe from the figure that disparity variation is less compared to our proposed real LF dataset, as shown in Figure \ref{fig: our_real_histo}.

\subsection{Proposed dataset description}

The proposed LF image dataset consists of LF images captured using Lytro Illum. While capturing the dataset from Lytro Illum, we have considered the proper focal point to maximize the variation of disparity with depth. Synthetic images were rendered using Blender software, and during the rendering process, we ensured that all viewpoints lay on a plane to maintain the inverse depth and disparity relationship.

\begin{figure}
    \centering
    \includegraphics[width=7.3cm, height = 2.4cm, , trim=0cm 4cm 0cm 0cm, clip]{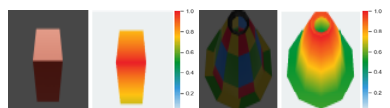}
    \vspace{-10pt}
\caption{Synthetic scenes in a single camera set-up rendered using Blender. On the left is the central sub-aperture image, and on the right is the corresponding normalized depth map generated using Blender.}
\label{synthetic1}
\vspace{-10pt}
\end{figure}

\begin{figure}[!htp]
    \centering
    \includegraphics[width=7.3cm,trim=0cm 5.6cm 0cm 0cm, clip]{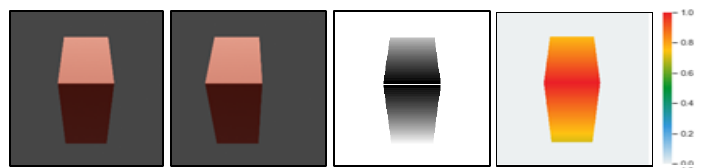}
    \vspace{-10pt}
\caption{Synthetic scenes in a stereo camera set-up rendered using Blender.Left to right:- Central view of the left Camera,  central view of the right camera, ground truth depth from Blender, and normalized depth map.}
\label{stereosynthetic}
\vspace{-10pt}
\end{figure}

\subsubsection{Real Scenes}

A real scene captured using Lytro Illum consists of an LFR (Light-Field Raw) file format as provided by the Lytro Illum camera. This contains 10 bits of uncompressed lenslet image data before demosaicing, in little-endian format. Furthermore, our dataset 
also contains 4D LF images, which were extracted from LFR files by using Light-Field Matlab Toolbox~\cite{danserumatlab}. Dimensions of a 4D LF image are $15\times15\times434\times625\times4$, where $15\times15$ represents the number of angular views, $434\times625$ represents the spatial resolution of each view, and $4$ corresponds to the RGB channels and weighting image component. We also provided the calibration data of the Lytro Illum camera. This calibration data consists of sensor metadata and white-field images specific to this camera and is used to process LFR files for extracting the 4D light-field data. The LF image dataset comprises $285$ LF images, captured in both indoor and outdoor conditions.

\paragraph{Single Camera}
The samples from our captured images in a single camera setup are shown in Figure~\ref{camerasetup}. For a few scenes, we have provided foreground and depth maps as shown in Figure~\ref{realScene1}. For the quantitative evaluation of the depth estimation algorithm, we also created an experimental setup as shown in Figure~\ref{camerasetup}. In this setup, we have placed different planner objects on a flex grid to ensure an exact distance between these objects and the LF camera. In this experimental setup for depth estimation, we captured $35$ images from different scenes with varying lighting conditions and camera ISO settings. This information is provided in the dataset. The focal length of the main lens was set to 30 mm, and the focal position was adjusted accordingly. The hyperfocal distance of the camera is 100 cm.

\paragraph{Stereo Camera}
We have captured stereo LF images using a light-field camera mounted on a mechanical gantry setup, as shown in Figure~\ref{camerasetup}, to simulate different baseline stereo light-field camera systems. We have collected data at a stereo baseline of 5 cm. The focal length of the main lens was kept fixed at 30mm. Different planner objects are kept at various distances, ranging from $20$ cm to $240$ cm, as shown in Figure~\ref{realStereo}. We have considered large occlusion of objects where one end of the object is visible, and in some cases, it was completely occluded in the right view. We have also provided ground truth distances measured from the camera and a foreground map. In this experimental setup for depth estimation, we captured $31$ stereo images from different scenes with varying lighting conditions and camera ISO settings. 

\subsubsection{Synthetic Scenes}

To obtain an appropriate dataset with intrinsic characteristics of the linear shift of virtual viewpoints in a two-dimensional plane, we have rendered multiple viewpoint images such that all the viewpoints lie on a plane with equal baselines among adjacent views. We use Blender software for rendering purposes. Each viewpoint image has a spatial resolution of $768\times768$. We have rendered 81 ($9\times9$) such viewpoint images to have a 4D LF of dimension $9\times9\times768\times768\times3$. $3$ corresponds to the RGB channels. In the dataset, we have provided a Blender source file, a 4D LF \textit{.mat} file for each LF image. For quantitative evaluation, the ground truth depth of each LF image is extracted from Blender.
\begin{table}[!htp]
\caption{Quantitative comparison of light-field depth estimation methods on single camera synthetic scenes.}
\vspace{-12pt}
\label{qtable1}
\center
\scalebox{0.7}{
\begin{tabular}{|c|c|c|c|c|c|c|c|c|}
\hline
Scene                                                                   & Cuboid & Cube  & Sphere & Geometric & Cone   & \makecell{Cone and\\ cylinder} & Floor 1 & Floor 2 \\ \hline
\cite{jeon2015accurate}  & 0.6225                           & 0.1666                           & 0.3519                           & 0.5457                              & 0.0577                           & 0.4053                                      & 0.0861                            & 0.0960                            \\ \hline
\cite{zhang2016robust}                            & 0.3360                           & 0.2702                           & 0.3235                           & 0.0488                              & 0.2273                           & 0.0970                                      & 0.3061                            & 0.1774                            \\ \hline
\cite{wang2015occlusion} & 0.0337                           & 0.1651                           & 0.0524                           & 0.1062                              & 0.0452                           & 0.0731                                      & 0.0645                            & 0.0617                            \\ \hline
\cite{park2017robust}  & 0.2782                           & 0.2444                           & 0.0650                           & 0.1380                              & 0.0411                           & 0.2142                                      & 0.1250                            & 0.0622                            \\ \hline
\cite{shin2018epinet}      & 0.6101                           & 0.3965                           & 0.3546                           & 0.4221                              & 0.0242                           & 0.2567                                      & 0.1572                            & 0.1881                            \\ \hline
\cite{mishiba2020}                                   & 0.1982                           & 0.1499                           & 0.0386                           & 0.1426                              & 0.0810                           & 0.0525                                      & 0.1978                            & 0.1678                            \\ \hline

\cite{nehra2021disparity}                                                                   & 0.0026 & 0.0028 & 0.0031 & 0.0202   & 0.0345 & 0.0184           & 0.0044  & 0.0342                            \\ \hline
\end{tabular}
}
\vspace{-12pt}
\end{table}
\begin{table}[!htp]
\caption{Quantitative comparison of light-field depth estimation methods on single camera real scenes.}
\vspace{-12pt}
\label{qtable2}
\center
\scalebox{0.7}{
\begin{tabular}{|c|c|c|c|c|}
\hline
Scene        & Image 1     & Image 2                              & Image 3                     & Image 4                     \\ \hline
Jeon et al. \cite{jeon2015accurate}                       & 0.4886                      & 0.052                                & 0.0081                      & 0.1997                      \\ \hline
Zhang et al. \cite{zhang2016robust} & 0.1302 & 0.0459 & 0.1159 & 0.1278 \\ \hline
Wang et al.  \cite{wang2015occlusion}                     & 0.2814                      & 0.1543                               & 0.231                       & 0.2504                      \\ \hline
Park et al. \cite{williem2016robust}                        & 0.2788                      & 0.1791                               & 0.1375                      & 0.2118                      \\ \hline
Shin \cite{shin2018epinet}                               & 0.1189                      & 0.1604                               & 0.1273                      & 0.1937                      \\ \hline
Mishiba  \cite{mishiba2020}                           & 0.1520                      & 0.1533                               & 0.3251                      & 0.2847                      \\ \hline
Nehra \textit{et al.} \cite{nehra2021disparity}                                                                & 0.0293            & 0.0554                               & 0.0066          & 0.0160            \\ \hline
\end{tabular}
}
\vspace{-12pt}
\end{table}

\paragraph{Single Camera}
The three-dimensional scenes are rendered by creating a two-dimensional grid of $9\times9$ viewpoints with a baseline of $3$ mm between adjacent views. The focal length is set to $50$ mm and has a spatial resolution of $768\times768$. We have rendered 9 scenes consisting of a variety of 3-D objects such as cube, cuboid, sphere, cone, and other geometrical shapes, as shown in Figure~\ref{synthetic1}.

\paragraph{Stereo Camera}
We have rendered $4$ scenes with two single-camera (left and right) LF images using the same experimental setup as the synthetic single camera for generating the synthetic stereo light-field dataset. The viewpoints of the left and right LF cameras lie on a plane. The baseline between the center viewpoint of the left camera and the center viewpoint of the right camera is $1$ meter. The focal length of the camera is $50$ mm. The synthetic image dataset contains occlusion, far-range images, repetitive patterns with curved surfaces, and complex geometry objects, as shown in Figure~\ref{stereosynthetic}.

\begin{table}[!htp]
\caption{Quantitative comparison of light-field depth estimation methods on stereo camera synthetic scenes.}
\vspace{-12pt}
\label{qtable3}
\centering
\scalebox{0.7}{
\begin{tabular}{|c|c|c|c|c|c|c|}
\hline
Image Name & Geometric & Cuboid & Cube  & Sphere & \makecell{Cone\\and cylinder} & \makecell{Geometric\\ occlusion} \\ \hline
Jeon \textit{et al.} \cite{jeon2015accurate}        & 0.35      & 0.29   & 0.34  & 0.51   & 0.59              & 0.54                \\ \hline
Wang \textit{et al.} \cite{wang2015occlusion}       & 0.42      & 0.27   & 0.05  & 0.1    & 0.072             & 0.49                \\ \hline
Zhang \textit{et al.} \cite{zhang2016robust}      & 0.12      & 0.17   & 0.17  & 0.15   & 0.1               & 0.15                \\ \hline
Nehra \textit{et al.}  \cite{nehra2021disparity}      & 0.019     & 0.053  & 0.006 & 0.057  & 0.031             & 0.038               \\ \hline
Yang \textit{et al.} \cite{yang2012non}        & 0.37      & 0.33   & 0.38  & 0.1    & 0.04              & 0.53                \\ \hline
Honsi \textit{et al.} \cite{hosni2012fast}      & 0.13      & 0.21   & 0.23  & 0.13   & 0.015             & 0.13                \\ \hline
Liu \textit{et al.} \cite{liu2019binocular}        & 0.035     & 0.21   & 0.27  & 0.12   & 0.029             & 0.031               \\ \hline
Yan \textit{et al.} \cite{yan2019depth}        & 0.031     & 0.19   & 0.22  & 0.016  & 0.032             & 0.011               \\ \hline
Nehra \textit{et al.} \cite{nehra2022depth}           & 0.012     & 0.021  & 0.001 & 0.024  & 0.023             & 0.008               \\ \hline
\end{tabular}
}
\vspace{-12pt}
\end{table}
\begin{table}[!htp]
\caption{Quantitative comparison of light-field depth estimation methods on stereo camera real scenes.}
\vspace{-12pt}
\label{qtable4}
\centering
\scalebox{0.7}{
\begin{tabular}{|c|c|c|c|c|c|c|c|c|}
\hline
Scene & 1 & 2 & 3 & 4 & 5 & 6 & 7 & 8 \\ \hline
Jeon \textit{et al.} \cite{jeon2015accurate}       & 0.17    & 0.22    & 0.14    & 0.18    & 0.15    & 0.15    & 0.12    & 0.12    \\ \hline
Wang \textit{et al.} \cite{wang2015occlusion}      & 0.043   & 0.19    & 0.21    & 0.18    & 0.18    & 0.19    & 0.14    & 0.095   \\ \hline
Zhang \textit{et al.} \cite{zhang2016robust}      & 0.37    & 0.32    & 0.4     & 0.36    & 0.47    & 0.41    & 0.34    & 0.35    \\ \hline
Nehra \textit{et al.}  \cite{nehra2021disparity}     & 0.026   & 0.02    & 0.01    & 0.052   & 0.033   & 0.035   & 0.039   & 0.023   \\ \hline
Yang \textit{et al.} \cite{yang2012non}       & 0.5     & 0.54    & 0.4     & 0.05    & 0.25    & 0.34    & 0.45    & 0.54    \\ \hline
Honsi \textit{et al.} \cite{hosni2012fast}     & 0.38    & 0.28    & 0.29    & 0.12    & 0.32    & 0.28    & 0.34    & 0.32    \\ \hline
Liu \textit{et al.} \cite{liu2019binocular}        & 0.24    & 0.14    & 0.014   & 0.018   & 0.011   & 0.013   & 0.063   & 0.064   \\ \hline
Yan \textit{et al.} \cite{yan2019depth}        & 0.19    & 0.12    & 0.019   & 0.33    & 0.018   & 0.023   & 0.098   & 0.041   \\ \hline
Nehra \textit{et al.} \cite{nehra2022depth}     & 0.01    & 0.019   & 0.011   & 0.003   & 0.011   & 0.01    & 0.015   & 0.011   \\ \hline
\end{tabular}
}
\vspace{-12pt}
\end{table}

\subsection{Depth Estimation Benchmark}

We have provided a quantitative light-field benchmarking on our proposed dataset. For quantitative analysis, we have used a normalized disparity map to have an independent comparison between the state-of-the-art methods on our proposed dataset. Normalized disparity value $N\_disp(i,j)$ at a pixel $(i,j)$ in terms of depth is defined as, 
\begin{equation}
N\_disp(i,j)= \frac{min\_depth}{depth(i,j)},
\label{nomralize_depth}
\end{equation} 
where $min\_depth$ is the minimum depth of a 3-D point in the entire scene, and $depth(i,j)$ is the depth of a 3-D point imaged at pixel location $(i,j)$. Similarly, in terms of disparity, normalized disparity is defined as,
\begin{equation}
N\_disp(i,j)= \frac{disp(i,j)}{max\_disp},
\end{equation} 
where $max\_disp$ is the maximum disparity of a 3-D point in the entire scene and $disp(i,j)$ is the disparity of a 3-D point imaged at pixel location $(i,j)$. 
The Mean Squared Error (MSE) between the normalized ground truth depth map $GT(i,j)$ and the estimated normalized depth map $N_{-}disp(i,j)$ is defined as, 
\begin{equation}
MSE = \frac{\sum\sum \left(GT(i,j)-N_{-}disp(i,j)\right)^2}{M},
\end{equation}
where $i$ and $j$ are pixel indices, such that $GT(i,j)>0$ and $M$ is the number of those pixels.

For quantitative comparison single camera dataset, we compared with the depth estimation methods proposed by Jeon \textit{et al.}~\cite{jeon2015accurate}, Zhang \textit{et al.}~\cite{zhang2016robust}, Wang \textit{et al.}~\cite{wang2015occlusion},  Williem \textit{et al.}~\cite{park2017robust}, Shin \textit{et al.}~\cite{shin2018epinet}, Mishiba~\cite{mishiba2020}, Nehra \textit{et al.}~\cite{nehra2021disparity}. Table~\ref{qtable1} shows the MSE between the normalized disparity computed from the ground truth and the normalized disparity estimated using various methods on single-camera synthetic scenes. Table~\ref{qtable2} shows the MSE between the normalized disparity computed from ground truth and the normalized disparity estimated using various methods on single-camera real scenes.

For quantitative comparison on stereo camera dataset, we compared with the depth estimation methods proposed by Jeon \textit{et al.}~\cite{jeon2015accurate}, Wang \textit{et al.}~\cite{wang2015occlusion}, Zhang \textit{et al.}~\cite{zhang2016robust}, Nehra \textit{et al.}~\cite{nehra2021disparity}, Yang \textit{et al.}~\cite{yang2012non}, Honsi \textit{et al.}~\cite{hosni2012fast}, Liu \textit{et al.}~\cite{liu2019binocular}, Yan \textit{et al}~\cite{yan2019depth}, and Nehra \textit{et al.}~\cite{nehra2022depth}. Table \ref{qtable3} shows the MSE between the normalized disparity computed from the ground truth and the normalized disparity estimated using various methods on the stereo camera synthetic scenes. Similarly, Table \ref{qtable4} shows MSE between normalized disparity computed from ground truth and normalized disparity estimated using various methods on stereo camera real scenes.

\section{Conclusion}

In this paper, we have introduced a real and synthetic light-field dataset for disparity-based light-field depth estimation. We also provided stereo light-field data for real scenes using a Lytro Illum camera with a mechanical gantry system, and synthetic stereo scenes were rendered using Blender. We also conducted an experimental study on the characteristics of a real micro-lens-based LF camera, analyzing the behavior of disparity with changes in depth and focal distance. We have highlighted the challenges present in existing LF datasets and demonstrated how our proposed dataset addresses these challenges. Our dataset exhibits sufficient disparity variation, making it suitable for disparity-based depth estimation. Unlike existing synthetic datasets, the introduced synthetic scenes in our dataset exhibit similar disparity characteristics to those of a real LF camera. We also provided ground-truth distances of objects and the foreground map in a scene to perform quantitative analysis of depth estimation.

\bibliographystyle{ACM-Reference-Format}
\bibliography{mainbib}

\appendix

\end{document}